\documentclass[sn-mathphys,Numbered]{sn-jnl}

\usepackage{graphicx}%
\usepackage{multirow}%
\usepackage{amsmath,amssymb,amsfonts}%
\usepackage{amsthm}%
\usepackage{mathrsfs}%
\usepackage[title]{appendix}%
\usepackage{xcolor}%
\usepackage{textcomp}%
\usepackage{manyfoot}%
\usepackage{booktabs}%
\usepackage{algorithm}%
\usepackage{algorithmicx}%
\usepackage{algpseudocode}%
\usepackage{listings}%

\begin{document}

\title[Article Title]{NEOLAF, an LLM-powered neural-symbolic cognitive architecture}


\author*[1,4,5]{\fnm{Richard Jiarui} \sur{Tong}}\email{richard.tong@ieee.org}

\author*[2,5]{\fnm{Cassie Chen} \sur{Cao}}\email{ccao5@sheffield.ac.uk}

\author[3]{\fnm{Timothy Xueqian} \sur{Lee}}\email{xql2001@tc.columbia.edu}


\author{\fnm{Guodong} \sur{Zhao}}\email{gzhao1997@gmail.com}

\author{\fnm{Ray} \sur{Wan}}

\author[4]{\fnm{Feiyue} \sur{Wang}}

\author[6]{\fnm{Xiangen} \sur{Hu}}\email{xhu@memphis.edu}

\author[7]{\fnm{Robin} \sur{Schmucker}}\email{rschmuck@cs.cmu.edu}

\author[8]{\fnm{Jinsheng} \sur{Pan}}\email{jpan24@ur.rochester.edu}

\author[9]{\fnm{Julian} \sur{Quevedo}}\email{julianhquevedo@gmail.com}

\author[10]{\fnm{Yu} \sur{Lu}}\email{luyu@bnu.edu.cn}


\affil[1]{\orgdiv{IEEE Artificial Intelligent Standards Committee}, \orgname{IEEE}, \orgaddress{ \state{NJ}, \country{USA}}}

\affil[2]{\orgname{University of Sheffield}, \orgaddress{\state{Sheffield}, \country{UK}}}

\affil[3]{\orgname{National Institute of Education}, \orgaddress{\country{Singapore}}}

\affil[4]{\orgname{Macao University of Science and Technology},  \orgaddress{\country{Macao}}}

\affil[5]{\orgname{Carnegie Learning}, \orgaddress{\state{PA}, \country{USA}}}

\affil[6]{\orgname{University of Memphis}, \orgaddress{\state{TN}, \country{USA}}}

\affil[7]{\orgname{Carnegie Mellon University}, \orgaddress{\state{PA}, \country{USA}}}

\affil[8]{\orgname{University of Rochester}, \orgaddress{\state{NY}, \country{USA}}}

\affil[9]{\orgname{Stanford University}, \orgaddress{\state{CA}, \country{USA}}}

\affil[10]{\orgname{Beijing Normal University}, \orgaddress{\state{Beijing}, \country{China}}}


\abstract{This paper presents the Never Ending Open Learning Adaptive Framework (NEOLAF), an integrated neural-symbolic cognitive architecture that models and constructs intelligent agents. The NEOLAF framework is a superior approach to constructing intelligent agents than both the pure connectionist and pure symbolic approaches due to its explainability, incremental learning, efficiency, collaborative and distributed learning, human-in-the-loop enablement, and self-improvement. The paper further presents a compelling experiment where a NEOLAF agent, built as a problem-solving agent, is fed with complex math problems from the open-source MATH dataset. The results demonstrate NEOLAF's superior learning capability and its potential to revolutionize the field of cognitive architectures and self-improving adaptive instructional systems.}

\keywords{cognitive architecture, agent, LLM, neural-symbolic}

\maketitle

\section{Purpose}\label{sec1}

This paper presents the Never Ending Open Learning Adaptive Framework (NEOLAF), which is an integrated neural-symbolic cognitive architecture. It can be used to model and construct intelligent agents, such as self-improving intelligent tutor agents in an adaptive instructional system environment. The design is inspired by the human cognitive development process, especially the human learning process. The authors propose that the NEOLAF framework is a superior approach to constructing intelligent agents than both the pure connectionist approach, such as ChatGPT and the Yann Lecun Approach \cite{LeCun2022} and the pure symbolic approach such as SOAR \cite{Laird2012TheSC} and ACT-R \cite{Ritter2018ACTRAC}, for several reasons, including its explainability, incremental learning, efficiency, collaborative and distributed learning, human-in-the-loop enablement\cite{10109264}, avoidance of hallucinations\cite{maynez2020faithfulness}, and self-improvement.

\section{Methodology}\label{sec2}
\subsection{NEOLAF Agent}\label{sec2.1}

NEOLAF learning agents are instantiated from a DNA-like starter kit (the "innate or system-0 layer") leveraging pre-trained large language models (LLMs) as the foundation for their reasoning and cognition. The agent employs and develops its cognitive capabilities at two levels akin to human cognition. The first is a system-1 “fast” process relying on zero-shot LLM response, and the second a “slow” system-2 process employing explicit reasoning and external services.\cite{booch2020thinking} NEOLAF agents utilize natural language as the primary communication interface and are self-learning, developing new cognitive capabilities by acquiring and applying grounded knowledge gradually through experience.

\subsection{KSTAR framework}\label{sec2.2}


KSTAR (knowledge, situation, task, action, result) is both a representation of the knowledge-experience duality and a process to acquire knowledge through experience by planning, doing, and learning in a continuous, recursive, and multitasking manner.

The KSTAR representation breaks down any encounter into its constituent parts: Situation, Task, Action, and Result:

\begin{enumerate}
\item Identify the \textbf{Situation}: Start by describing the specific situation or context in which the experience occurred.
\item Define the \textbf{Task}: Determine the objective or goal that needs to be accomplished within the given situation. In KSTAR, the task may be further decomposed into subtasks and the implied co-tasks of planning, forecasting, and grounding.
\item Plan and Execute the \textbf{Actions}: Identify the specific actions taken to address the task at hand. Describe the steps, decisions, or strategies employed to accomplish the goal. The actions can be further represented by the subject-verb-objectives structure (agent-skill-constraints) to describe both the planned and actual action steps.
\item Forecast and Evaluate the \textbf{Result}: Forecast the outcome or result of the actions taken as part of the plan and assess whether the objective of the main task and co-tasks have been successfully achieved by grounding to truth and reality and consider the consequences, impact, and any feedback or reactions received as a result of the actions.
\end{enumerate}

The NEOLAF agent uses LLM capability to reason from both explicit \textbf{Knowledge} accumulated incrementally from prior encounters and in context learning as well as implicit knowledge encoded in LLM to guide its interactions. Knowledge reflects comparisons of expected results given prior knowledge, the situation, task, and planned actions, against actual outcomes after the action is executed.

\begin{itemize}
\item Expected Result: Before taking any action, NEOLAF agents generate an anticipated outcome based on prior knowledge, the situation, and task.
\item Planned Action: NEOLAF agents formulate a plan or strategy to achieve the expected result. This involves making decisions, setting goals, and determining the steps to be taken.
\item Actual Result: After executing the planned action, NEOLAF agents observe, evaluate, and encode the outcome that actually occurs.
\end{itemize}

\subsection{Memory}\label{sec2.3}


Memory in NEOLAF agents are reflected in two ways.

\begin{itemize}

\item Explicit memory: Encoded prior knowledge reflecting agents’ KSTAR process for each encounter.

\item Implicit Memory: Offline knowledge injection is used for model fine-tuning and knowledge embedding plug-ins to improve the system-1 zero-shot performance. This is analogous to memory consolidation during human sleep.

\end{itemize}

\subsection{Related Work}\label{sec2.4}

Our approach utilizes \textit{chain of thought} reasoning to conduct planning and learning tasks in KSTAR process using recent advances in LLM for reasoning\cite{lightman2023lets} \cite{lewkowycz2022solving}  \cite{wang2022self}  \cite{wei2022chain}  \cite{nye2021show}  \cite{kojima2022large}, RL\cite{sutton2023alberta}, multitask learning\cite{shen2021generate}, and planning\cite{fabiano2023fast}. 

\section{Result}\label{sec3}

\subsection{Initial Implementation}\label{sec3.1}

While the NEOLAF architecture is intended to serve as a domain-general architecture, initial implementation and evaluation are ongoing for a math problem-solving agent. The NEOLAF agent will be instantiated and trained on math problems from the open-source MATH dataset. We hypothesise that the NEOLAF agent’s continuous learning process will allow it to develop reasoning capabilities that will enable it to perform well on problems beyond its training set and achieve problem-solving capabilities rivalling leading LLMs (e.g. GPT-4).

In our initial experiment, we will evaluate the performance of four different AI setups for performance: 1) ChatGPT 3.5, 2) ChatGPT 4.0, 3) A ‘Chain of Thought’ model with WolframAlpha and other plugins, and 4) a NEOLAF Agent, on metrics such as answer correctness, response time, and explanation quality. 50 test cases from AIME and USAMO Math Competitions will be utilized as a test set. These challenging problems would require novel approaches that diverge substantially from the simpler problems found in the MATH dataset, providing a challenging test of the capabilities hypothesized to emerge from the NEOLAF continuous learning process.

\subsection{Application to Education}\label{sec3.2}


We intend for NEOLAF to serve as a cognitive architecture underlying functions of an agent-based learning environment (Open Learning Adaptive Framework - OLAF) involving 3 types of agents - learners, human teachers, and AI agents. OLAF will combine a learning management approach from XAPI CMI5 to allow teacher and student interaction and using an autonomous agent framework similar to the Stanford Interactive Simulacra\cite{Park2023GenerativeAI}.

\section{Conclusions}\label{sec4}

The key elements of the NEOLAF architecture include the integration of system-1 LLM capabilities with system-2 explicit reasoning and external services, problem-solving within encounters using the KSTAR representation, and learning and memory through encoding prior experiences using the KSTAR representation and model fine-tuning.

We contend that the combination of these elements, drawing inspiration from human cognition and combining symbolic and connectionist approaches to AI, will overcome some of the key challenges of either approach. We expect that NEOLAF agents, powered by a local LLM, will display highly competitive performance in a diverse array of tasks through continual learning from experience. This would constitute a lightweight and continually improving AI model, improving on current leading LLMs that are expensive to train and maintain.

Beyond the math problem-solving experiment and learning settings in OLAF, we hope to introduce multimodal reasoning\cite{gupta2022visual} to the agent framework and build a co-habitat ecosystem called BotLand, designed as a complex adaptive multi-agent environment where both intelligent agents and humans can interact and evolve socially. We hypothesize that the potential of using complex adaptive systems (CAS) and evolution to achieve artificial generalized intelligence (AGI) is better than the alternative routes.

\begin{appendices}






\end{appendices}


\bibliography{sn-bibliography}


\begin{thebibliography}{17}
\ifx \bisbn   \undefined \def \bisbn  #1{ISBN #1}\fi
\ifx \binits  \undefined \def \binits#1{#1}\fi
\ifx \bauthor  \undefined \def \bauthor#1{#1}\fi
\ifx \batitle  \undefined \def \batitle#1{#1}\fi
\ifx \bjtitle  \undefined \def \bjtitle#1{#1}\fi
\ifx \bvolume  \undefined \def \bvolume#1{\textbf{#1}}\fi
\ifx \byear  \undefined \def \byear#1{#1}\fi
\ifx \bissue  \undefined \def \bissue#1{#1}\fi
\ifx \bfpage  \undefined \def \bfpage#1{#1}\fi
\ifx \blpage  \undefined \def \blpage #1{#1}\fi
\ifx \burl  \undefined \def \burl#1{\textsf{#1}}\fi
\ifx \doiurl  \undefined \def \doiurl#1{\url{https://doi.org/#1}}\fi
\ifx \betal  \undefined \def \betal{\textit{et al.}}\fi
\ifx \binstitute  \undefined \def \binstitute#1{#1}\fi
\ifx \binstitutionaled  \undefined \def \binstitutionaled#1{#1}\fi
\ifx \bctitle  \undefined \def \bctitle#1{#1}\fi
\ifx \beditor  \undefined \def \beditor#1{#1}\fi
\ifx \bpublisher  \undefined \def \bpublisher#1{#1}\fi
\ifx \bbtitle  \undefined \def \bbtitle#1{#1}\fi
\ifx \bedition  \undefined \def \bedition#1{#1}\fi
\ifx \bseriesno  \undefined \def \bseriesno#1{#1}\fi
\ifx \blocation  \undefined \def \blocation#1{#1}\fi
\ifx \bsertitle  \undefined \def \bsertitle#1{#1}\fi
\ifx \bsnm \undefined \def \bsnm#1{#1}\fi
\ifx \bsuffix \undefined \def \bsuffix#1{#1}\fi
\ifx \bparticle \undefined \def \bparticle#1{#1}\fi
\ifx \barticle \undefined \def \barticle#1{#1}\fi
\bibcommenthead
\ifx \bconfdate \undefined \def \bconfdate #1{#1}\fi
\ifx \botherref \undefined \def \botherref #1{#1}\fi
\ifx \url \undefined \def \url#1{\textsf{#1}}\fi
\ifx \bchapter \undefined \def \bchapter#1{#1}\fi
\ifx \bbook \undefined \def \bbook#1{#1}\fi
\ifx \bcomment \undefined \def \bcomment#1{#1}\fi
\ifx \oauthor \undefined \def \oauthor#1{#1}\fi
\ifx \citeauthoryear \undefined \def \citeauthoryear#1{#1}\fi
\ifx \endbibitem  \undefined \def \endbibitem {}\fi
\ifx \bconflocation  \undefined \def \bconflocation#1{#1}\fi
\ifx \arxivurl  \undefined \def \arxivurl#1{\textsf{#1}}\fi
\csname PreBibitemsHook\endcsname

\bibitem[\protect\citeauthoryear{LeCun}{2022}]{LeCun2022}
\begin{botherref}
\oauthor{\bsnm{LeCun}, \binits{Y.}}:
A path towards autonomous machine intelligence.
Meta - Fundamental AI Research
(2022).
Version 0.9.2
\end{botherref}
\endbibitem

\bibitem[\protect\citeauthoryear{Laird}{2012}]{Laird2012TheSC}
\begin{bchapter}
\bauthor{\bsnm{Laird}, \binits{J.E.}}:
\bctitle{The soar cognitive architecture}.
(\byear{2012})
\end{bchapter}
\endbibitem

\bibitem[\protect\citeauthoryear{Ritter et~al.}{2018}]{Ritter2018ACTRAC}
\begin{barticle}
\bauthor{\bsnm{Ritter}, \binits{F.E.}},
\bauthor{\bsnm{Tehranchi}, \binits{F.}},
\bauthor{\bsnm{Oury}, \binits{J.D.}}:
\batitle{Act-r: A cognitive architecture for modeling cognition.}
\bjtitle{Wiley interdisciplinary reviews. Cognitive science}
\bvolume{10 3},
\bfpage{1488}
(\byear{2018})
\end{barticle}
\endbibitem

\bibitem[\protect\citeauthoryear{Tong and Lee}{2023}]{10109264}
\begin{barticle}
\bauthor{\bsnm{Tong}, \binits{J.}},
\bauthor{\bsnm{Lee}, \binits{T.}}:
\batitle{Trustworthy ai that engages humans as partners in teaching and
  learning}.
\bjtitle{Computer}
\bvolume{56}(\bissue{05}),
\bfpage{62}--\blpage{73}
(\byear{2023})
\doiurl{10.1109/MC.2023.3234517}
\end{barticle}
\endbibitem

\bibitem[\protect\citeauthoryear{Maynez et~al.}{2020}]{maynez2020faithfulness}
\begin{botherref}
\oauthor{\bsnm{Maynez}, \binits{J.}},
\oauthor{\bsnm{Narayan}, \binits{S.}},
\oauthor{\bsnm{Bohnet}, \binits{B.}},
\oauthor{\bsnm{McDonald}, \binits{R.}}:
On faithfulness and factuality in abstractive summarization.
arXiv preprint arXiv:2005.00661
(2020)
\end{botherref}
\endbibitem

\bibitem[\protect\citeauthoryear{Booch et~al.}{2020}]{booch2020thinking}
\begin{botherref}
\oauthor{\bsnm{Booch}, \binits{G.}},
\oauthor{\bsnm{Fabiano}, \binits{F.}},
\oauthor{\bsnm{Horesh}, \binits{L.}},
\oauthor{\bsnm{Kate}, \binits{K.}},
\oauthor{\bsnm{Lenchner}, \binits{J.}},
\oauthor{\bsnm{Linck}, \binits{N.}},
\oauthor{\bsnm{Loreggia}, \binits{A.}},
\oauthor{\bsnm{Murugesan}, \binits{K.}},
\oauthor{\bsnm{Mattei}, \binits{N.}},
\oauthor{\bsnm{Rossi}, \binits{F.}},
\oauthor{\bsnm{Srivastava}, \binits{B.}}:
Thinking Fast and Slow in AI
(2020)
\end{botherref}
\endbibitem

\bibitem[\protect\citeauthoryear{Lightman et~al.}{2023}]{lightman2023lets}
\begin{botherref}
\oauthor{\bsnm{Lightman}, \binits{H.}},
\oauthor{\bsnm{Kosaraju}, \binits{V.}},
\oauthor{\bsnm{Burda}, \binits{Y.}},
\oauthor{\bsnm{Edwards}, \binits{H.}},
\oauthor{\bsnm{Baker}, \binits{B.}},
\oauthor{\bsnm{Lee}, \binits{T.}},
\oauthor{\bsnm{Leike}, \binits{J.}},
\oauthor{\bsnm{Schulman}, \binits{J.}},
\oauthor{\bsnm{Sutskever}, \binits{I.}},
\oauthor{\bsnm{Cobbe}, \binits{K.}}:
Let's Verify Step by Step
(2023)
\end{botherref}
\endbibitem

\bibitem[\protect\citeauthoryear{Lewkowycz et~al.}{2022}]{lewkowycz2022solving}
\begin{botherref}
\oauthor{\bsnm{Lewkowycz}, \binits{A.}},
\oauthor{\bsnm{Andreassen}, \binits{A.}},
\oauthor{\bsnm{Dohan}, \binits{D.}},
\oauthor{\bsnm{Dyer}, \binits{E.}},
\oauthor{\bsnm{Michalewski}, \binits{H.}},
\oauthor{\bsnm{Ramasesh}, \binits{V.}},
\oauthor{\bsnm{Slone}, \binits{A.}},
\oauthor{\bsnm{Anil}, \binits{C.}},
\oauthor{\bsnm{Schlag}, \binits{I.}},
\oauthor{\bsnm{Gutman-Solo}, \binits{T.}}, et al.:
Solving quantitative reasoning problems with language models.
arXiv preprint arXiv:2206.14858
(2022)
\end{botherref}
\endbibitem

\bibitem[\protect\citeauthoryear{Wang et~al.}{2022}]{wang2022self}
\begin{botherref}
\oauthor{\bsnm{Wang}, \binits{X.}},
\oauthor{\bsnm{Wei}, \binits{J.}},
\oauthor{\bsnm{Schuurmans}, \binits{D.}},
\oauthor{\bsnm{Le}, \binits{Q.}},
\oauthor{\bsnm{Chi}, \binits{E.}},
\oauthor{\bsnm{Zhou}, \binits{D.}}:
Self-consistency improves chain of thought reasoning in language models.
arXiv preprint arXiv:2203.11171
(2022)
\end{botherref}
\endbibitem

\bibitem[\protect\citeauthoryear{Wei et~al.}{2022}]{wei2022chain}
\begin{botherref}
\oauthor{\bsnm{Wei}, \binits{J.}},
\oauthor{\bsnm{Wang}, \binits{X.}},
\oauthor{\bsnm{Schuurmans}, \binits{D.}},
\oauthor{\bsnm{Bosma}, \binits{M.}},
\oauthor{\bsnm{Chi}, \binits{E.}},
\oauthor{\bsnm{Le}, \binits{Q.}},
\oauthor{\bsnm{Zhou}, \binits{D.}}:
Chain of thought prompting elicits reasoning in large language models.
arXiv preprint arXiv:2201.11903
(2022)
\end{botherref}
\endbibitem

\bibitem[\protect\citeauthoryear{Nye et~al.}{2021}]{nye2021show}
\begin{botherref}
\oauthor{\bsnm{Nye}, \binits{M.}},
\oauthor{\bsnm{Andreassen}, \binits{A.J.}},
\oauthor{\bsnm{Gur-Ari}, \binits{G.}},
\oauthor{\bsnm{Michalewski}, \binits{H.}},
\oauthor{\bsnm{Austin}, \binits{J.}},
\oauthor{\bsnm{Bieber}, \binits{D.}},
\oauthor{\bsnm{Dohan}, \binits{D.}},
\oauthor{\bsnm{Lewkowycz}, \binits{A.}},
\oauthor{\bsnm{Bosma}, \binits{M.}},
\oauthor{\bsnm{Luan}, \binits{D.}}, et al.:
Show your work: Scratchpads for intermediate computation with language models.
arXiv preprint arXiv:2112.00114
(2021)
\end{botherref}
\endbibitem

\bibitem[\protect\citeauthoryear{Kojima et~al.}{2022}]{kojima2022large}
\begin{botherref}
\oauthor{\bsnm{Kojima}, \binits{T.}},
\oauthor{\bsnm{Gu}, \binits{S.S.}},
\oauthor{\bsnm{Reid}, \binits{M.}},
\oauthor{\bsnm{Matsuo}, \binits{Y.}},
\oauthor{\bsnm{Iwasawa}, \binits{Y.}}:
Large language models are zero-shot reasoners.
arXiv preprint arXiv:2205.11916
(2022)
\end{botherref}
\endbibitem

\bibitem[\protect\citeauthoryear{Sutton et~al.}{2023}]{sutton2023alberta}
\begin{botherref}
\oauthor{\bsnm{Sutton}, \binits{R.S.}},
\oauthor{\bsnm{Bowling}, \binits{M.}},
\oauthor{\bsnm{Pilarski}, \binits{P.M.}}:
The Alberta Plan for AI Research
(2023)
\end{botherref}
\endbibitem

\bibitem[\protect\citeauthoryear{Shen et~al.}{2021}]{shen2021generate}
\begin{botherref}
\oauthor{\bsnm{Shen}, \binits{J.}},
\oauthor{\bsnm{Yin}, \binits{Y.}},
\oauthor{\bsnm{Li}, \binits{L.}},
\oauthor{\bsnm{Shang}, \binits{L.}},
\oauthor{\bsnm{Jiang}, \binits{X.}},
\oauthor{\bsnm{Zhang}, \binits{M.}},
\oauthor{\bsnm{Liu}, \binits{Q.}}:
Generate \& rank: A multi-task framework for math word problems.
arXiv preprint arXiv:2109.03034
(2021)
\end{botherref}
\endbibitem

\bibitem[\protect\citeauthoryear{Fabiano et~al.}{2023}]{fabiano2023fast}
\begin{botherref}
\oauthor{\bsnm{Fabiano}, \binits{F.}},
\oauthor{\bsnm{Pallagani}, \binits{V.}},
\oauthor{\bsnm{Ganapini}, \binits{M.B.}},
\oauthor{\bsnm{Horesh}, \binits{L.}},
\oauthor{\bsnm{Loreggia}, \binits{A.}},
\oauthor{\bsnm{Murugesan}, \binits{K.}},
\oauthor{\bsnm{Rossi}, \binits{F.}},
\oauthor{\bsnm{Srivastava}, \binits{B.}}:
Fast and Slow Planning
(2023)
\end{botherref}
\endbibitem

\bibitem[\protect\citeauthoryear{Park et~al.}{2023}]{Park2023GenerativeAI}
\begin{botherref}
\oauthor{\bsnm{Park}, \binits{J.S.}},
\oauthor{\bsnm{O'Brien}, \binits{J.C.}},
\oauthor{\bsnm{Cai}, \binits{C.J.}},
\oauthor{\bsnm{Morris}, \binits{M.R.}},
\oauthor{\bsnm{Liang}, \binits{P.}},
\oauthor{\bsnm{Bernstein}, \binits{M.S.}}:
Generative agents: Interactive simulacra of human behavior.
ArXiv
\textbf{abs/2304.03442}
(2023)
\end{botherref}
\endbibitem

\bibitem[\protect\citeauthoryear{Gupta and Kembhavi}{2022}]{gupta2022visual}
\begin{botherref}
\oauthor{\bsnm{Gupta}, \binits{T.}},
\oauthor{\bsnm{Kembhavi}, \binits{A.}}:
Visual Programming: Compositional visual reasoning without training
(2022)
\end{botherref}
\endbibitem

\end{thebibliography}

\end{document}